\documentclass[letterpaper, 10 pt, conference]{ieeeconf}  

\IEEEoverridecommandlockouts                              

\usepackage{graphics} 
\usepackage{graphicx}
\usepackage{epsfig} 
\usepackage{amsmath} 
\usepackage{amssymb}  
 \usepackage{algorithm}
\usepackage[compatible]{algpseudocode}
\algnewcommand\AAND{\textbf{ and }}
\algnewcommand\Or{\textbf{ or }}
\usepackage{color}
\usepackage{citesort}
\usepackage{flushend}
\usepackage{url}

\DeclareMathAlphabet{\pazocal}{OMS}{zplm}{m}{n}

\newcommand{\Ws}{\pazocal{W}}

\newcommand{\Cs}{\pazocal{C}}

\newcommand{\Is}{\pazocal{I}}

\usepackage{array}
\newcolumntype{C}[1]{>{\centering\arraybackslash}p{#1}}
\newcolumntype{M}[1]{>{\raggedright\arraybackslash}p{#1}}

\usepackage{array} 
\newcolumntype{L}[1]{>{\raggedright\let\newline\\\arraybackslash\hspace{0pt}}m{#1}}	
\newcolumntype{S}[1]{>{\centering\let\newline\\\arraybackslash\hspace{0pt}}m{#1}}
\newcolumntype{R}[1]{>{\raggedleft\let\newline\\\arraybackslash\hspace{0pt}}m{#1}}

\makeatletter
\renewcommand*{\@opargbegintheorem}[3]{\trivlist
 \item[\hskip \labelsep{\bfseries #1\ #2}] \textbf{(#3)}\ }
\makeatother

\title{\LARGE \bf Keyframe-based Direct Thermal-Inertial Odometry}

\author{Shehryar Khattak, Christos Papachristos, and Kostas Alexis
\thanks{This material is based upon work related to the Mine Inspection Robotics project sponsored by the Nevada Knowledge Fund administered by the Governor's Office of Economic Development.}
\thanks{This material is based upon work supported by the Defense Advanced Research Projects Agency (DARPA) under Agreement No. HR00111820045. The presented content and ideas are solely those of the authors.}
\thanks{The authors are with the Autonomous Robots Lab, University of Nevada, Reno, 1664 N. Virginia, 89557, Reno, NV, USA
        {\tt\small shehryar.khattak@nevada.unr.edu}}%
}

\begin{document}

\maketitle
\thispagestyle{empty}
\pagestyle{empty}

\begin{abstract}
This paper proposes an approach for fusing direct radiometric data from a thermal camera with inertial measurements to extend the robotic capabilities of aerial robots for navigation in GPS--denied and visually degraded environments in the conditions of darkness and in the presence of airborne obscurants such as dust, fog and smoke. An optimization based approach is developed that jointly minimizes the re--projection error of \textbf{$\textrm{3D}$} landmarks and inertial measurement errors. The developed solution is extensively verified against both ground--truth in an indoor laboratory setting, as well as inside an underground mine under severely visually degraded conditions. 
\end{abstract}

\section{INTRODUCTION}\label{sec:intro}
Aerial robots have recently seen an increased utilization in a wide variety of tasks typically reserved for humans as their flexibility makes them suitable for a diverse set of applications, while also mitigating risk to human life and lowering costs~\cite{roboticinspectionsurvey,RHEM_ICRA_2017,changeDetetion,bircher2018receding,m3u,NasaUavElectrical2017,visionDepth,karrer2016real,ICUAS2017,ROSChapter}. An important aspect of their application is towards performing critical tasks in GPS--denied, poorly--illuminated and sensor--degraded environments, such as underground mines and tunnels. To navigate GPS--denied environments aerial robots rely on their on--board sensing to estimate their pose. Traditionally RGB cameras have been the sensor of choice due to their low weight, cost and power requirements. However, in poorly--illuminated conditions and in the presence of airborne obscurants, such as dust, fog and smoke, the image quality of RGB cameras degrades significantly making them unreliable for pose estimation. In contrast thermal cameras, e.g. Long Wave Infrared (LWIR) sensors, are not affected by scene illumination changes or by the presence of certain obscurants~\cite{ThermalPerception}, making them a viable sensing alternative.

\begin{figure}[h!]
\centering
    \includegraphics[width=0.99\columnwidth]{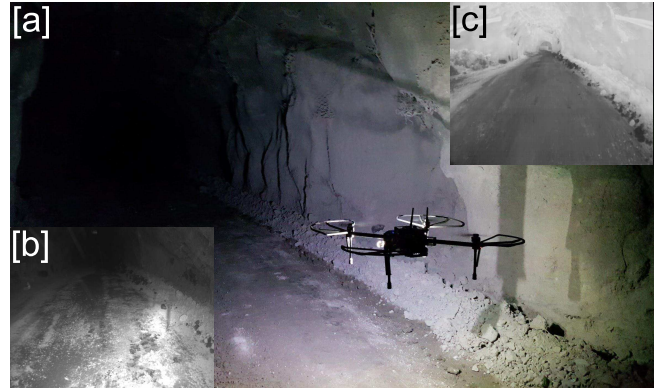}
\caption{An instance of the underground mine experiment. [a] shows the thermal camera equipped aerial robot navigating through the mine shaft. [b] shows an image from a visible light camera on--board the aerial robot with shutter--synced LED illumination. Not only the visible light camera cannot observe the environment at a distance but it is also obstructed by the airborne dust. [c] shows the thermal image of the same scene providing visibility at a longer range while not being affected by presence of obscurants.}\label{fig:main}
\end{figure}

However, previous approaches that utilized thermal cameras for robot pose estimation present a set of limitations. First, most previously proposed solutions have operated on re--scaled thermal images rather than utilizing full radiometric information provided by thermal cameras~\cite{selectiveVisualThermal2013,visionThermal,mutlispectral2015,rgbtSLAM2017,thermalMarker}. The motivation behind this choice relates to the fact that thermal cameras typically capture LWIR data in more than 8--bit resolution (e.g., 14--bit) but feature matching methods for visual images require the data to be re--scaled to 8--bit resolution. However, re--scaling of LWIR data results in loss of information observed in the form of lower contrast, resulting in poor feature matching performance, as shown in~\cite{thermalstereo2015}. As a workaround, these approaches use histogram equalization techniques, usually implemented as Automatic Gain Control (AGC), to improve contrast of LWIR images and by extension improving feature matching performance. However, as AGC adapts the image according to the range of LWIR spectrum currently present in the scene, the image contrast can change significantly if hot or cold objects enter the Field--of--View (FoV) of the thermal camera. This is especially problematic for a moving camera, such as one on--board an aerial robot, as the observed view of the environment is dynamic. Another approach is to disable AGC completely and set an acceptable range of thermal information to be re--scaled into 8-bit resolution~\cite{ICUAS2018Thermal}. However, this requires manually setting a range for operation in a particular environment under certain thermal conditions and is also not suitable for long term navigation, because as thermal cameras operate, they accumulate sensor noise. Hence, if only a small range of thermal information is scaled to 8--bit information then the effect of noise can also be amplified, resulting in image artifacts. Thermal cameras reduce this noise accumulation by performing a periodic Flat Field Correction (FFC), during which camera operation is suspended for up to $500$ milliseconds and a uniform temperature (flat field) is presented to the camera sensor for the estimation of noise correction parameters. Application of FFC operation in itself presents two challenges. First, if only a small range of thermal information is re--scaled, the difference in image intensity values will be very different after a correction is applied. Such a large change in intensity is problematic for both direct and feature--based methods to establish robust correspondences between images.
Secondly, the interruption of image data can make odometry estimates prone to drift, especially in filter--based solutions such as~\cite{rovio} where state estimates are progressively propagated. However, this is less of a problem for optimization--based odometry approaches that operate over a temporal window and state estimate is only propagated when new data is available and optimized against a temporal history. 


Motivated by the discussion above, in this work we present a key--frame based odometry estimation approach that uses direct 14--bit radiometric data from a monocular LWIR thermal camera to establish correspondences between successive images. Working directly on 14--bit radiometric data allows our approach to overcome many of the problems associated with image re--scaling, to operate without relying on feature detection and description methods designed for visual systems, and to remain generalizable to a variety of environments under different thermal conditions. Similarly, our approach is able to remain robust against data interruption due to its optimization based estimation nature. Furthermore, we integrate measurements from an Inertial Measurement Unit (IMU) to formulate a joint cost function for our thermal--inertial odometry estimation approach. The motivation behind integrating inertial measurements is threefold. First, IMU measurements provide a transformation prior for the image alignment process. Second, they provide direct observation of two rotation states reducing the number of unobserved pose degrees of freedom from $6$ to $4$. Third, they provide a better estimation of scale in the case of monocular vision. To verify our proposed solution, a set of experiments are conducted including, a) comparison of odometry estimation of an aerial robot, in complete darkness, against ground truth provided by a VICON system, as well as, b) odometry estimation of an aerial robot traversing through an underground mine in dark and dust-filled conditions. An instance of this experiment is shown in Figure~\ref{fig:main}, and a video of the conducted experiments and the derived results can be found at~\url{https://youtu.be/-hnL5kLqT4Q}.

The remainder of the paper is structured as follows: Section:~\ref{sec:approach} details the proposed approach, followed by the experimental evaluation results presented in Section:~\ref{sec:experiments}. Finally, conclusions are drawn in Section:~\ref{sec:concl}.

\section{PROPOSED APPROACH}\label{sec:approach}
Our proposed approach of fusing direct 14--bit radiometric data with inertial measurements for odometry estimation can be divided into a front--end component and a back--end component. This bifurcation allows us to run our odometry estimation framework in a multi--threaded manner on a modern CPU. The key responsibility of the front--end component is to perform an alignment between an incoming image to previous images in the camera coordinate frame $(\Cs)$ based on the minimization of radiometric error and to initialize $3\textrm{D}$ landmarks in the world coordinate frame $(\Ws)$. Given a set of $3\textrm{D}$ landmarks, the key responsibility of the back--end component is to estimate odometry by jointly minimizing the re--projection errors in landmark positions and the intra--frame inertial measurement errors over a sliding window. An overview of the approach is shown in Figure~\ref{fig:system_overview}. Each component is detailed below:

\begin{figure*}[h!]
\centering
    \includegraphics[width=\textwidth]{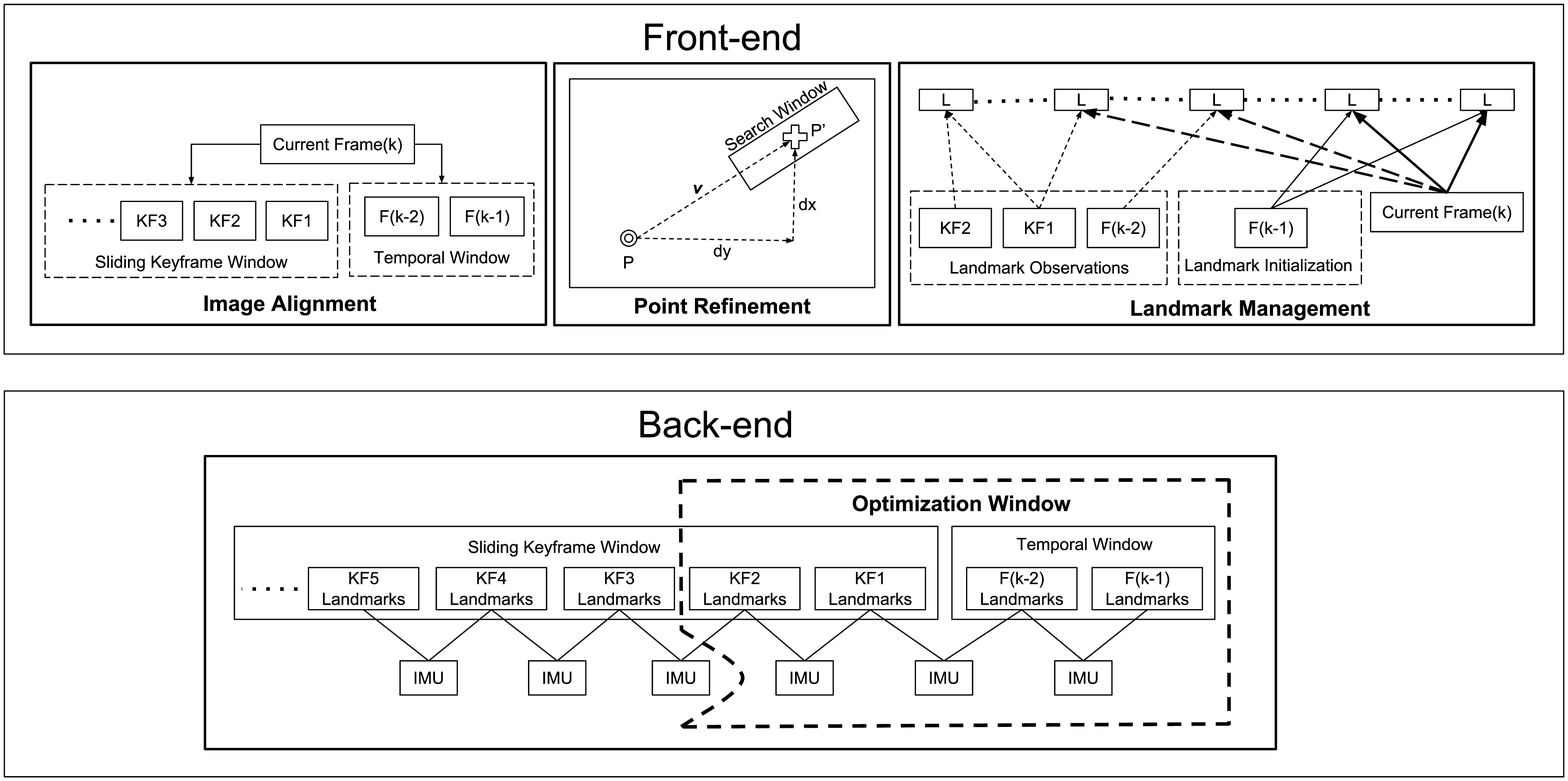}
\caption{An overview of the proposed approach. The front--end component is responsible for aligning the incoming image to the last two received images and to the last two key--frames. If alignment of the current image with a past images or key--frames is successful, location of points in the current image are refined to determine their localization quality. Points with good localization quality are then triangulated to either add observations to previously observed landmarks or initialize new landmarks. Using the initialized landmarks the back--end is responsible for jointly optimizing the landmark re--projection and intra--frame inertial measurement errors in a sliding window approach to estimate the robot pose.}\label{fig:system_overview}
\end{figure*}

\subsection{Front--End Design}

In our framework, the front--end component replaces all of the tasks usually associated with feature detection, matching and pruning in feature--based approaches. Our front--end component can be divided into four sub--components, namely: i) Image Alignment, ii) Point Initialization,  iii) Point Refinement, and iv) Landmark Initialization and Pruning. Each of these sub--components are discussed below.

\subsubsection{Image Alignment}

\begin{figure}[h!]
\centering
    \includegraphics[width=0.99\columnwidth]{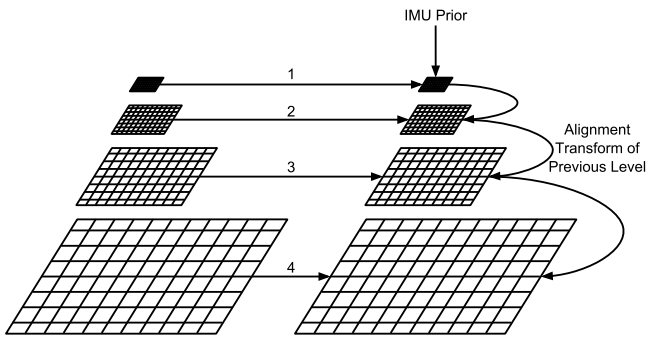}
\caption{Image alignment starts at the coarsest level of the image scale--space pyramid with an IMU prior. More iterations are allowed for the solution to converge at the higher levels of the pyramid. If alignment is successful the the transformation estimate becomes a prior for the alignment of the next level.}\label{fig:alignment}
\end{figure}

To align and establish correspondences between two images we minimize the radiometric error between them, making our approach belong to the category of direct approaches. Historically, direct visual odometry methods based on the minimization of photometric error such as~\cite{DTAM,LSDSLAM,REMODE}, require optimization to be performed on a large number of points and thus becoming computationally expensive. Inspired by~\cite{DSO}, we instead track a sparse set of points for performing image alignment. Given a set of points in a reference image, we project them into the incoming image as:

\vspace{-1.5ex}
\small
\begin{eqnarray}\label{eqs:reprojection_eq}
{p}'=K\left ( RK^{-1}\left ( p,d^{-1} \right ) + t \right )
\end{eqnarray}
\normalsize
where $p$ and ${p}'$ are the original and new projected point locations in pixels respectively, $K$ is the intrinsic camera matrix, $R$ and $t$ are the rotation matrix and translation vector  between frames respectively and $d$ is the estimated depth of the point. As $R$ and $t$ are unknown, we estimate them by minimizing the radiometric error between points. If the depth of a point is known from its previous association to a $3D$ landmark, it is used, otherwise $t$ is estimated only up to scale. In order to make our radiometric error minimization process more robust, we calculate the error over a small neighborhood around each point:

\vspace{-1.5ex}
\small
\begin{eqnarray}\label{eqs:residual_eq}
\mathbf{e_{radio}}=\sum_{i\epsilon P}^{ }\sum_{p\epsilon N_{i}}^{ }\|T\left({p}'\right)-T\left(p\right)\|_{2}
\end{eqnarray}
\normalsize
where $\mathbf{e_{radio}}$ is the squared sum of radiometric errors for a set of tracked points $(P)$ calculated over the neighborhood $(N_{i})$ of each point with $T$ representing the thermal value of each point in 14--bit resolution. A weighted Gauss--Newton optimization is then performed to estimate the transformation parameters between the two images. However, performing an optimization for a large number of points can be computationally expensive. Instead we perform our alignment operation over the levels of an image scale--space pyramid in a top--down manner, where the coarsest levels of the two images are aligned first and their transformation estimate becomes an alignment prior for the next levels. This allows the solution to converge at the lowest level of the image pyramid in very few iterations. Furthermore, we start the alignment process at the coarsest level with a transformation prior provided by the IMU, as shown in Figure~\ref{fig:alignment}. Once the transformation parameters between the frames are estimated, points in the reference image are projected into the new image and are considered matched points. These matched points are then used for image alignment for the next image. During the alignment process we align the incoming image to the last two incoming images and the latest two key--frames. The last two incoming images form a small sliding temporal window whereas the latest two key--frames are outside of this window and further in the past.

\subsubsection{Point Initialization}
To choose points for tracking we first calculate image gradients over the whole image. We then divide the image into $32\times32$ pixel blocks and calculate the median gradient $(g_{med})$ in each block. A constant offset $(g_{off})$ is added to the $g_{med}$ of each block and all gradient values in the block below this $g_{med}+g_{off}$ value are suppressed. To ensure a good distribution of points is chosen across the scene, we start by picking the points with largest non-suppressed gradient in each block and insert them into a $2\textrm{D}$ point selection grid equal in size to the image. This $2\textrm{D}$ point selection grid allows us to quickly check if the next candidate point has sufficient distance from the previously selected points without explicitly calculating a distance metric. This approach ensures that the points selected have sufficient gradient values and are well distributed across the image. We select points in this manner over the image scale--space such that the selected number of points at a given level are half in number to those in the level below it. New points are initialized in an image only when the number of tracked points falls below a certain threshold. In such a case the successfully tracked points are first inserted into the $2\textrm{D}$ grid before inserting new points. Each time new points are initialized we set that image as a key--frame.

\subsubsection{Point Refinement}
Once a reference and a new image are aligned and pixel locations of points in the new image are determined, these pixel locations are then further refined to understand how well a point can be localized in its neighborhood. We calculate the localization quality of a point by calculating its radiometric residual along the vector of its motion in a search window centered around the point as shown in Figure~\ref{fig:system_overview}. We then compute the ratio of the lowest residual to the second lowest residual in the search window and reject the point if this ratio is lower than a threshold. This step ensures that points that are sufficiently constrained in their location are allowed to progress, while eliminating points that due to alignment errors separate from corner or edge locations on to a uniform planar surface and become unconstrained in their location.

\subsubsection{Landmark Initialization and Pruning}
Once point correspondences between images have been established, we attempt to initialize them as $3\textrm{D}$ landmarks. We use the OpenGV~\cite{opengv} library to triangulate points and check if rays emanating from points are not parallel, indicating points that are far away. If a point can be successfully triangulated, it is assigned a landmark ID. New landmarks are only initialized when image alignment is successful between current frame and the latest frame in the temporal window. However, for each successful alignment of the current frame with the oldest frame in the temporal window and the key--frames, new observations are added to the already initialized landmarks. If enough landmarks are observable between two frames, we check for the quality of landmarks by using their depth estimates to back--project their pixel locations from one frame to the other and compare the back--projected pixel location to the one obtained after the image alignment and point refinement process. Landmarks with inconsistent depth have larger error in their pixel locations and can be pruned out. The remaining landmarks, if having a minimum number of observations, are added to the optimization back--end for odometry estimation.

\subsection{Back-End Design}

Given a set of $3\textrm{D}$ landmarks and inertial measurements between image frames we estimate the pose of the robot by solving a non--linear optimization problem that minimizes the re--projection errors of the observed landmarks, while respecting inertial constraints. Our back--end design takes inspiration from~\cite{okvis} and the re--projection error of a landmark can be written as:

\vspace{-1.5ex}
\small
\begin{eqnarray}\label{eqs:reproj_eq}
\mathbf{e_{reproj}}=p_{\Cs}-K\left(T_{\Cs\Is}T_{\Is\Ws}l_{\Ws}\right)
\end{eqnarray}
\normalsize
where $p_{\Cs}$ is the point coordinates in $\Cs$, $l_{\Ws}$ is the corresponding $3D$ landmark location in $\Ws$, $T_{\Cs\Is}$ is the transformation from IMU in the IMU frame $\Is$ to $\Cs$ and $T_{\Is\Ws}$ is the transformation of the IMU in $\Ws$. Furthermore, to write the inertial measurement error model we first define our IMU state equations similarly to the ones described in~\cite{rovio,okvis}:

\vspace{-1.5ex}
\small
\begin{eqnarray}\label{eqs:imu_state_eq}
\mathbf{x} = [\mathbf{r}~\mathbf{q}~\mathbf{v}~\mathbf{b}_a~\mathbf{b}_\omega]^{'}
\end{eqnarray}
\normalsize
where $\mathbf{r}$ and $\mathbf{q}$ are the position and orientation of the IMU in $\Ws$ respectively, $\mathbf{v}$ is the velocity of the IMU in $\Is$, $\mathbf{b}_a~$ and $\mathbf{b}_\omega~$ are the estimated bias of the accelerometer and gyroscope expressed in $\Is$. The differential equations for the state elements are defined as:

\vspace{-1.5ex}
\small
\begin{eqnarray}\label{eqs:diff_eq}
\begin{aligned}
&\dot{\mathbf{r}} = R_{\Ws\Is}\mathbf{v}
\\
&\dot{\mathbf{q}} = -q(\mathbf{\hat{\omega}})
\\
&\dot{\mathbf{v}} = -\hat{\omega^{\times}}\mathbf{v} + \mathbf{\hat{a}} + R_{\Is\Ws}\mathbf{g}
\\
&\dot{\mathbf{b}_{a}} = w_{a}
\\
&\dot{\mathbf{b}_{\omega}} = w_{\omega}
\end{aligned}
\end{eqnarray}
\normalsize
where $R_{\Ws\Is}$ is the rotation matrix $\Is\xrightarrow{}\Ws$, $^{\times}$ represents skew-symmetric matrix of a vector, $\mathbf{g}$ is the gravity vector in $\Ws$, $w_\star$ are white Gaussian noise processes, while $\mathbf{\hat{a}}$ and $\mathbf{\hat{\omega}}$ represent the bias corrected proper acceleration and angular velocity and are given as:

\vspace{-1.5ex}
\small
\begin{eqnarray}\label{eqs:reproj_eq}
\begin{aligned}
&\mathbf{\hat{a}} = \tilde{a} - b_{a} + w_{a}
\\
&\mathbf{\hat{\omega}} = \tilde{\omega} - b_{\omega} + w_{\omega}
\end{aligned}
\end{eqnarray}
\normalsize
where $ \tilde{a}$ and $\tilde{\omega}$ are the uncompensated IMU measurements. Therefore, the error term for inertial measurements between two frames can be written as:

\vspace{-1.5ex}
\small
\begin{eqnarray}\label{eqs:reproj_eq}
\mathbf{e_{imu}} = \mathbf{\hat{x}}^{k+1} - \mathbf{x}^{k}
\end{eqnarray}
\normalsize
where $\mathbf{x}^{k}$ represents the IMU state at the acquisition time of frame $k$ and $\mathbf{\hat{x}}^{k+1}$ represents the predicted state of the IMU at the acquisition time of frame $k+1$. A forward Euler integration scheme is used to calculated the predicted IMU state. Given the re--projection and inertial measurement error equations, the cost function for the joint thermal--inertial problem can be written as:

\vspace{-1.5ex}
\small
\begin{eqnarray}\label{eqs:reproj_eq}
J = \sum_{k=1}^{K} \sum_{l\epsilon L(k)}^{ } \mathbf{e_{reproj}^{'k,l,}} W^{k,l}_{reproj} \mathbf{e_{reproj}^{k,l}}
+\sum_{k=1}^{K}\mathbf{e_{imu}^{'k}} W^{k}_{imu} \mathbf{e_{imu}^{k}}
\end{eqnarray}
\normalsize
where $k$ represents the frame being processed in the sliding window $K$ containing temporal frames and key-frames, $L(k)$ represents the set of landmarks observable in frame $k$, $\mathbf{e_{reproj}}$ represents the stacked vector of re--projection errors of every landmark $l$ visible in frame $k$, $W^{k,l}_{reproj}$ represents the co--variance matrix for landmark measurements in frame $k$, and $W^{k}_{imu}$ represents the co--variance matrix for IMU measurements. This cost function is then minimized using the Google Ceres~\cite{ceres} optimization framework and produces an estimate of the robot pose. 

\section{EXPERIMENTAL EVALUATION}\label{sec:experiments}
To evaluate the performance of our thermal--inertial odometry estimation approach, a set of experiments were performed both indoors and in an underground mine using an aerial robot in challenging conditions of poor--illumination and in the presence of airborne obscurants. A video detailing the experiments can be found at~\url{https://youtu.be/-hnL5kLqT4Q}

\subsection{System Overview}
For the purpose of experimental studies a DJI Matrice 100 quad--rotor was used. An Intel NUC Core--i7 computer (NUC7i7BNH) was carried on--board the robot. A FLIR Tau 2 thermal camera was mounted on the robot to provide thermal images of $640\times512$ resolution at $30$ frames per second. The intrinsic calibration parameters of the thermal camera were calculated using our custom designed thermal checker board pattern~\cite{ICUAS2018Thermal}. To provide inertial measurements a VN--100 MEMS IMU from VectorNav was employed. The camera--to--IMU extrinsics were identified based on the work in~\cite{kalibr}.

\subsection{Indoor Ground Truth Comparison}
\begin{figure}[h!]
\centering
    \includegraphics[width=0.99\columnwidth]{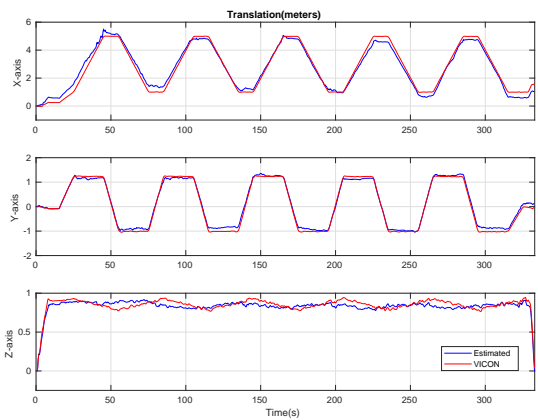}
\caption{Plots show translation along each axis as the aerial robot followed the predefined trajectory. Translation estimates are compared to ground truth provided by a VICON system.}\label{fig:vicon_translation_plot}
\end{figure}

\begin{figure}[h!]
\centering
    \includegraphics[width=0.99\columnwidth]{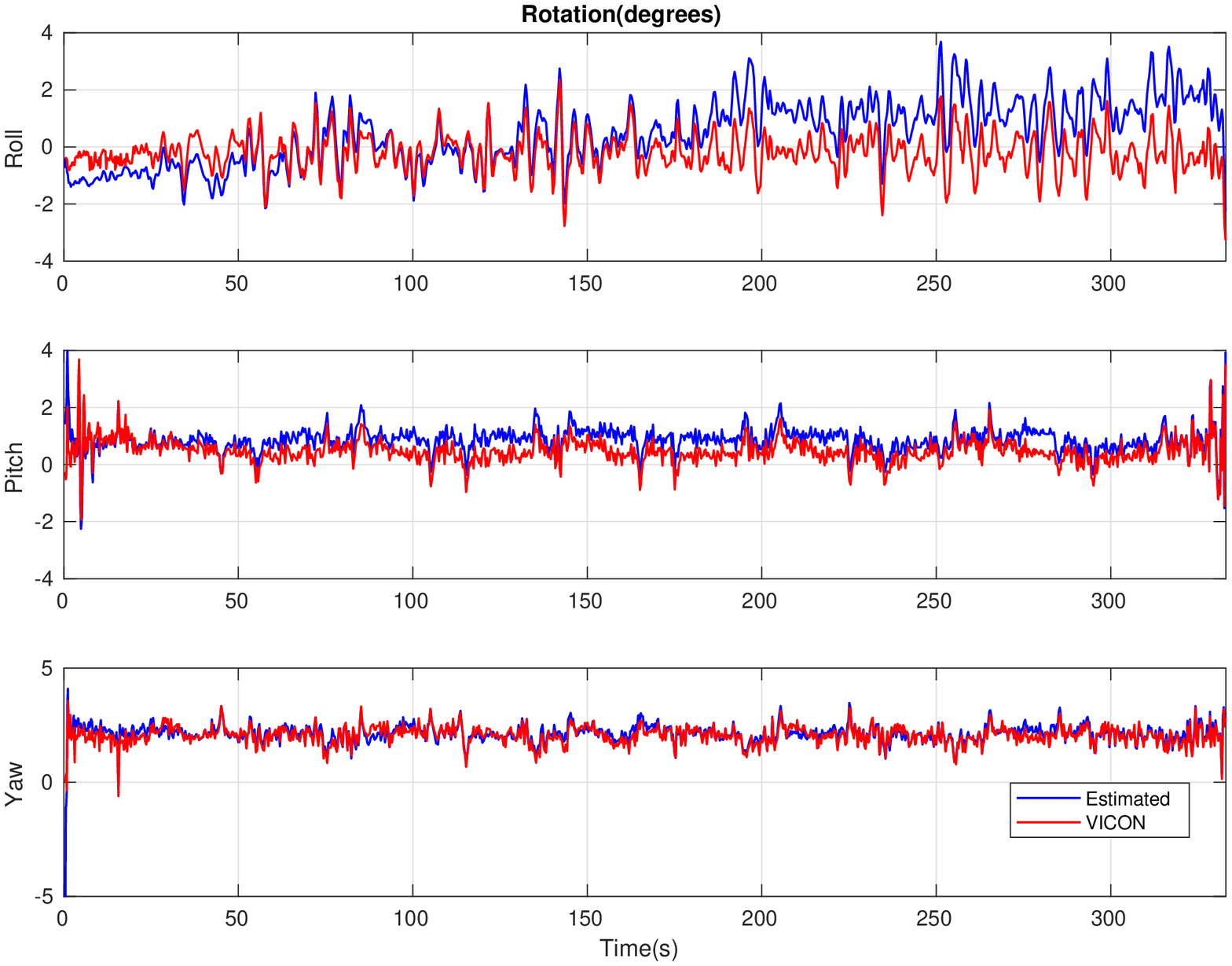}
\caption{Plots show rotation along each axis as the aerial robot followed the predefined trajectory. Rotation estimates are compared to ground truth provided by a VICON system.}\label{fig:vicon_rotation_plot}
\end{figure}

To evaluate the performance of our odometry solution, an experimental flight was performed in an indoor environment. To simulate the typical thermal profile of an indoor setting a computer, a room heater and some wires are introduced into the scene. A prescribed rectangular trajectory of $\textrm{length}=4.0\textrm{m}$ and $\textrm{width}=2.5\textrm{m}$ was executed 5 times in complete darkness. The estimated trajectory is compared against ground--truth provided by a VICON motion capture system. Figure~\ref{fig:vicon_translation_plot} and Figure~\ref{fig:vicon_rotation_plot} present the derived translation and orientation results for each axis with the proposed method. The root mean squared error over the full flight including landing and take--off are presented in Table:~\ref{tab:rmse}. It can be noted that the overall RMSE error is small.

\vspace{-0.2cm}
\begin{table}[h!]
\begin{center}
\caption[RMSE Pose Estimation]{RMSE in Pose Estimation with respect to VICON}
\begin{tabular}{|c|c|c|ll}
\cline{1-3}
\multicolumn{3}{|c|}{\textbf{RMSE in Pose Estimation}}              &  &  \\ \cline{1-3}
\textbf{Axis} & \textbf{Translation(m)} & \textbf{Rotation(deg)} &  &  \\ \cline{1-3}
\textbf{X}    & 0.2741                  & 1.1822                 &  &  \\ \cline{1-3}
\textbf{Y}    & 0.0910                  & 0.6380                 &  &  \\ \cline{1-3}
\textbf{Z}    & 0.0481                  & 0.4701                 &  &  \\ \cline{1-3}
\end{tabular}

\label{tab:rmse}
\end{center}
\end{table}
\vspace{-0.2cm}

\subsection{Comparison to other methods}
To provide a thorough evaluation of our method, we compare the performance of our approach against three state--of--the--art visual and visual--inertial odometry estimation methods namely, OKVIS~\cite{okvis}, DSO~\cite{DSO} and ROVIO~\cite{rovio}, on thermal imagery instead of other approaches that use thermal data for odometry estimation as mentioned in Section~\ref{sec:intro}. This choice was made primarily for two reasons 1) None of the approaches that utilize thermal data for odometry estimation work purely on thermal imagery like our approach and instead require both visual and thermal data 2) These approaches do not make use of full radiometric information but instead work on re--scaled images making them comparable to visual odometry approaches. Therefore it is more reasonable to compare our proposed method directly against state--of--art methods. 
\begin{figure}[h!]
\centering
    \includegraphics[width=0.99\columnwidth]{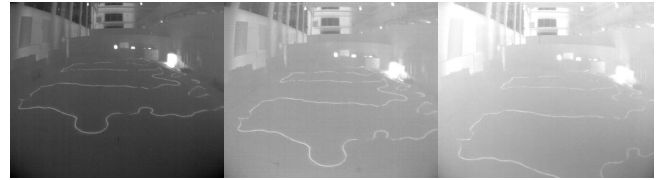}
\caption{An instance from the indoor flight experiment. The significance of noise accumulation can be seen in terms of change in contrast in the re--scaled thermal images. From left to right the image contrast changes significantly in a short period of time. After FFC is applied the contrast changes instantly from the right most image to the left most image.}\label{fig:degrade}
\end{figure}

\begin{figure}[h!]
\centering
\includegraphics[width=0.99\columnwidth,height=0.3\textheight]{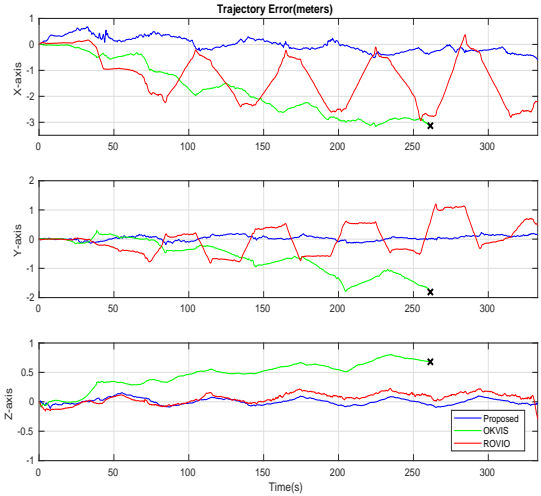}
\caption{Plots present trajectory errors with respect to ground--truth provided by a VICON system for our proposed approach, OKVIS and ROVIO. DSO trajectory was excluded from the plots due to multiple failures. The $\times$ marks the point in time where OKVIS odometry estimation failed. It can be noted our proposed approach has significantly smaller error in X and Y axis as compared to other approaches.}\label{fig:odometry_error_comparison}
\end{figure}

For odometry estimation comparison with state--of--the--art methods, we re--scale thermal images from the indoor flight data--set. As the thermal camera faces the same environment constantly, instead of using AGC we use pre--defined thresholds to re--scale the images to provide better contrast. However, to be fair to these approaches, the image quality is much lower as compared to the quality of visual data. The quality is even further degraded due to accumulation of noise as shown in Figure~\ref{fig:degrade}, where a sharp change in contrast can be noted. Due to low image quality, all of the evaluated approaches perform poorly on such image data causing their pose estimates to diverge significantly. By trial--and--error we try to find the best parameters for each approach, however even with significant fine tuning the odometry estimates of these approaches diverge significantly as shown in Figure~\ref{fig:odometry_error_comparison}.
To provide context to the results shown in Figure~\ref{fig:odometry_error_comparison}, we discuss the strengths and weaknesses of each of these state--of--the--art methods when operating on low contrast data. OKVIS, is a key--frame based optimization approach that relies on the detection and matching of BRISK~\cite{brisk} features. These features perform poorly on such low contrast data and provide very few matches. Lowering of feature detection threshold forces the detection of more features, however, matching performance of remains low resulting in unreliable odometry estimation. However, it should be noted that being an optimization--based approach, OKVIS was able to work through data interruption during the application of FFC operations.
ROVIO is a semi--direct filter--based method which relies on IMU data for its state propagation. ROVIO relies on image data to apply correction to its propagated state estimate. Delays in image data, as in case of thermal images, causes the state estimate to drift. As a filter--based method, ROVIO is not able to correct for drift errors. However, operating as semi--direct approach, ROVIO is able to find and track features even in low contrast images. DSO is a sparse direct odometry approach that relies on minimizing photometric error between images using a windowed optimization. Being a direct method DSO is sensitive to sudden intensity changes and constantly lost tracking during testing when operating on thermal images and hence was removed from plot in Figure~\ref{fig:odometry_error_comparison}. Continuous tracking loss forced a number re--initializations, which in turn lead to poor estimation of scale. DSO solely relies on image data for scale estimation unlike visual--inertial methods such as the method presented, OKVIS, and ROVIO that take advantage of inertial measurements for better estimation of scale. However, as DSO works on direct image data it was able to find many point correspondences even in low contrast images. This comparison study verifies the choices made during the development of our approach.

\subsection{Underground Mine Experiment}
\begin{figure}[h!]
\centering
    \includegraphics[width=0.99\columnwidth,height=0.22\textheight]{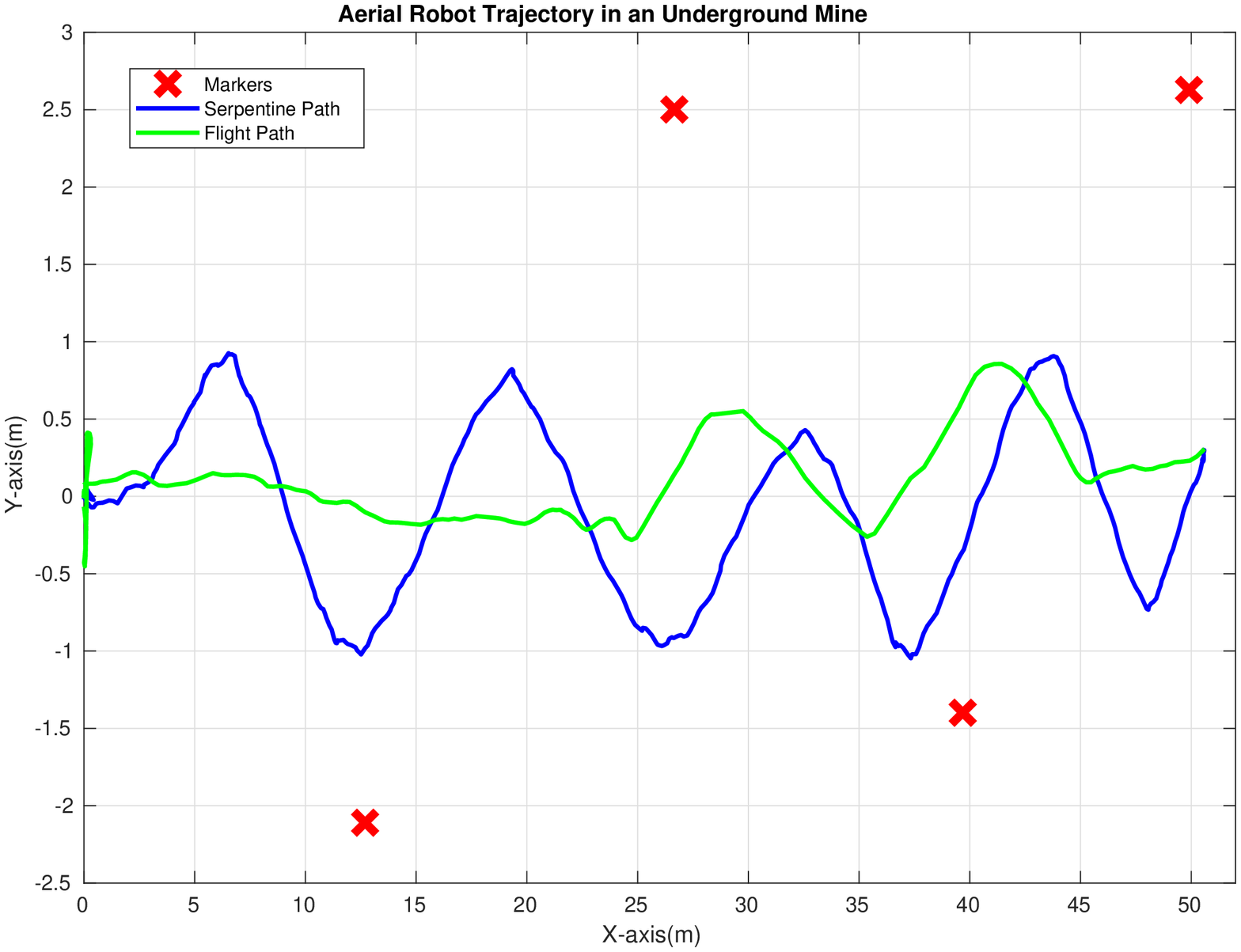}
\caption{Plot shows the robot odometry during the navigation of underground mine while following a serpentine path and a forward facing flight path as well as the locations of the placed markers.}\label{fig:mine_trajectory}
\end{figure}

To demonstrate the real world application and performance of our proposed method, we conducted tests in an underground mine in the conditions of darkness and in the presence of airborne dust. To provide ground--truth markers were placed along the mine shaft with the final marker position being $50.0\textrm{m}$ away from the take--off position. For verification a hand--held test was carried out by moving the robot along a serpentine path towards the final marker location. A flight test was then carried out with the robot facing towards the end of the mine shaft throughout its flight. Figure~\ref{fig:mine_trajectory} shows both of these paths as well as the marker locations. The error estimation in the direction of the mine shaft length is presented in Table:~\ref{tab:mine_error}. It can be seen that even in complete darkness and in the presence of airborne dust our approach was able to estimate robot pose with very low error.

\vspace{-0.2cm}
\begin{table}[h!]
\begin{center}
\caption[]{Trajectory error during the underground mine experiments}
\begin{tabular}{|c|c|c|}
\hline
                  & \textbf{Serpentine Path} & \textbf{Flight Path} \\ \hline
\textbf{Error(m)} & 0.5350                   & 0.6001               \\ \hline
\end{tabular}

\label{tab:mine_error}
\end{center}
\end{table}
\vspace{-0.2cm}

\section{CONCLUSIONS}\label{sec:concl}
In this paper we presented an approach to estimate robot odometry using direct radiometric data from a thermal camera. Our approach fuses direct thermal data with inertial measurements in a windowed optimization manner to overcome problems associated with long term thermal camera operation. We demonstrate the accuracy of our approach by comparing our odometry estimates for an aerial robot against the ground truth as well as state--of--the--art visual and visual--inertial odometry methods. We show the application of our method by using it for the estimation of odometry of an aerial robot deployed in an underground mine in the conditions of darkness and in the presence of airborne dust. In the future, we would focus on extending our method to perform re--localization when tracking is lost in order to enable long term operation of aerial robots in dark, dirty and dangerous environments.









\bibliographystyle{IEEEtran}
\bibliography{ICRA2019}

\end{document}